\def\year{2022}\relax
\documentclass[letterpaper]{article} 
\usepackage{aaai22}  
\usepackage{times}  
\usepackage{helvet}  
\usepackage{courier}  
\usepackage[hyphens]{url}  
\usepackage{graphicx} 
\usepackage{natbib}  
\usepackage{caption} 
\usepackage[section]{placeins}
\usepackage{amsmath}
\usepackage[hang,flushmargin]{footmisc}
\newcommand\workshopnote[1]{\renewcommand\thefootnote{}\footnote{#1}}
\DeclareCaptionStyle{ruled}{labelfont=normalfont,labelsep=colon,strut=off} 
\usepackage[linesnumbered,noend,lined,ruled,vlined]{algorithm2e}

\title{ Distributional Data Augmentation Methods for Low Resource Language}
\author{
     Mosleh Mahamud,
     Zed Lee, 
     Isak Samsten
}
\affiliations{
     Department of Computer and Systems Sciences\\
     Borgarfjordsgatan 12, Kista, Sweden\\
    \{mosleh.mahamud,zed.lee,samsten\}@dsv.su.se
}

\usepackage{bibentry}

\begin{document}

\maketitle

\begin{abstract}

Text augmentation is a technique for constructing synthetic data from an under-resourced corpus to improve predictive performance. Synthetic data generation is common in numerous domains. However, recently text augmentation has emerged in natural language processing (NLP) to improve downstream tasks. One of the current state-of-the-art text augmentation techniques is easy data augmentation (EDA), which augments the training data by injecting and replacing synonyms and randomly permuting sentences. One major obstacle with EDA is the need for versatile and complete synonym dictionaries, which cannot be easily found in low-resource languages. To improve the utility of EDA, we propose two extensions, easy distributional data augmentation (EDDA) and type specific similar word replacement (TSSR), which uses semantic word context information and part-of-speech tags for word replacement and augmentation. In an extensive empirical evaluation, we show the utility of the proposed methods, measured by F1 score, on two representative datasets in Swedish as an example of a low-resource language. With the proposed methods, we show that augmented data improve classification performances in low-resource settings.
\end{abstract}

\workshopnote{Accepted to Workshop on Knowledge Augmented Methods for Natural Language Processing, in conjunction with AAAI 2023.}

\section{Introduction}

Augmentation is a technique to construct synthetic training data from available datasets. Various augmentation techniques have been used mainly in the computer vision field to improve machine learning models \cite{shorten2021text}, especially with huge deep learning models in the area. However, text augmentation has been growing recently, also being aligned with massive models that have come out nowadays \cite{bayer2021survey}. The two core reasons to use text augmentation are as follows: 1) some languages are in low-resource domains, thus it is hard to get enough data to train the model. 2) augmentation can be helpful to strengthen decision boundaries, leading to more robust classifiers or better uncertainty estimates so the model can be more familiar with the local space around examples \cite{bayer2021survey}. Unlike images, languages cannot be generalized or merged, meaning each language only has its own resources, while images can easily be merged regardless of topics and types. In this sense, text augmentation techniques can benefit low-resource languages such as Swedish, Kazakh, Tamil, Welsh, Upper Serbian, and many more \cite{10.1162/coli_a_00425}.

There have been a few text augmentation techniques, from the most straightforward one \cite{ebrahimi2017hotflip,kolomiyets2011model}, to complex ones using separate deep learning models \cite{wu2019conditional, croce-etal-2020-gan, malandrakis-etal-2019-controlled}. One of the easiest ways to apply text augmentation is with a technique called easy data augmentation (EDA). EDA has four main techniques to augment a sentence \cite{wei-zou-2019-eda} as follows: synonyms replace (SR), random Insertion (RI), random swap (RS), and random deletion (RD). While EDA can be regarded as a universal text augmentation technique that can be applied to any language. However, this may not always be true, as it is not truly universal in the sense of not being able to apply to different languages since it still depends on other language-dependent modules such as wordnet. Adapting EDA to low-resource languages may be even more challenging since some language dependencies cannot be easily solved. Therefore, this paper aims to provide a framework for modified EDA augmentation that can also easily be applied to low-resource languages. We show our framework for Swedish as an example of a low-resource language.

While the Swedish language is classified into the low-resource group, there have been a few text augmentation trials for the language. One of the earliest text augmentation works has been done on clinical text data in Swedish by merging various sources of text for named entity recognition (NER) tasks using different deep models \cite{berg-dalianis-2019-augmenting}. However, this paper has a limitation in that it only tests on one Swedish clinical dataset and the augmentation techniques used in the paper are domain-specific, thus it cannot be applied to every Swedish text. Moreover, a group of researchers has tried controlled text perturbation using three main perturbation methods: n-gram shift, clause shift, and random shift on Swedish text \cite{taktasheva-etal-2021-shaking}. However, this paper focuses only on evaluating deep models such as BERT and BART \cite{devlin-etal-2019-bert,lewis-etal-2020-bart} and investigates attention layers for each token to observe their behavior without discussing the effects of augmentation on the models' performances. They also do not disclose how the augmentation techniques are implemented, hindering the possibility of reproducing the technique.

To the best of our knowledge, no previous work has been found where EDA with neural adaptation is applied to the Swedish text. Regarding the inner workings of EDA, it is heavily dependent on wordnet synonym replacement. As aforementioned, there may not always be a comprehensive dictionary in every language, especially in low-resource languages. Therefore, we replace wordnet with the word2vec \cite{10.5555/2999792.2999959,10.2307/42636393} model to integrate within this augmentation framework, which becomes a data-driven approach to augmenting data, which we call \textbf{E}asy \textbf{D}istributional  \textbf{D}ata  \textbf{A}ugmentation (\textbf{EDDA}). We expect that this approach can greatly help low-resource languages without good quality dictionary data, such as wordnet, use EDA techniques with a trainable component.

Moreover, we also introduce how syntax information of words can also be used to augment data, which we call \textbf{T}ype \textbf{S}pecific \textbf{S}imilar word \textbf{R}eplacement (\textbf{TSSR}). This is due to randomness in EDDA may affect sentence sentiment \cite{10.1145/3366424.3383552, bayer2021survey,anaby2020not} by producing sentimentally dissimilar synthetic sentences; therefore, this is a directed approach to complement EDDA.

\smallskip
\noindent \textbf{Contibutions. } The main contributions of this paper can be summarized as follows:

\begin{itemize}
\item We adapt EDA-style augmentation techniques for low-resource languages by using distributional synonym replacement that does not require strong language-specific dependency. We exemplify its usefulness in Swedish text.
\item We introduce and evaluate a novel augmentation method using POS information, which we name TSSR, as a complementary module to our EDDA framework and show that this method can significantly improve predictive performance.
\item We show that by using the proposed augmentation techniques, we increase the F1 score only using 40\%-50\% of the training data compared to the baseline performances without augmentation.
\item We provide our code in the GitHub repository for reproducibility purposes\footnote{\url{https://github.com/mosh98/Text_Aug_Low_Res}}.

\end{itemize}

\section{Related work}

Among the multitude of text perturbation techniques, text augmentation comes down to two main categories: symbolic and neural augmentation techniques \cite{shorten2021text}. The first consists of a wide range of techniques, such as rule-based augmentation, feature-space augmentation, and graph-structured augmentation, whereas the latter is based on different techniques of deep neural networks, such as back-translation, style augmentation, and generative data augmentation. Symbolic augmentation is more interesting because it can be more controllable and interpretable than its counterpart. However, very little research has been done where symbolic and neural augmentation techniques are aligned to augment sentences which this paper explores. 

As text augmentation is a relatively new area, there have not been many experiments in the Swedish domain. As per our knowledge, the earliest attempts with augmentation in the Swedish language are with Swedish clinical text mining, where they merge various sources of text for NER \cite{berg-dalianis-2019-augmenting}. Apart from that, no popular augmentation techniques like EDA have been applied to the SuperLim suites of benchmarking datasets \cite{Adesam-Yvonne2020-299130}, which we showcase in this paper.

Swedish has been known to be a low-resource language \cite{10.1162/coli_a_00425}. Hence, there is a need for available resources such as EDA or other augmentation tools that could improve various NLP downstream tasks. One paper discusses pretraining for an ASR model in low-resource domains where they have tried various augmentation techniques \cite{DBLP:journals/corr/abs-1910-10762}. However, it focuses on augmentation for speech data and is not universal or applicable to purely text models.

One of the first attempts at EDA in Swedish can be found in an unreliable news detection problem \cite{munoz-sanchez-etal-2022-first}. This paper deals with a classification problem where three main augmentation techniques have been applied to boost the model's performance, such as (1) sub-sampling of data, (2) EDA, and (3) back translation. Both back translation and EDA are also combined to achieve good classification performance. They train with a bag of words model, Bi-LSTM, and BERT in their experiments. The paper denotes that EDA functions best with simple machine learning models. However, this is where this paper and our work diverge, as (1) we use a neural-adapted EDA that can easily adapt to any language. (2) we focus on testing our methods on two benchmarking datasets. Our paper takes inspiration from EDA and infuses it with a word2vec making it a data-driven approach to text perturbation.

Similar augmentation attempts have been made on the DALAJ dataset \cite{volodina-etal-2021-dalaj} with controlled perturbations using three main perturbation methods: N-gram shift, clause shift, and random sift. N-gram shifts are about utilizing compound nouns and prepositions to perturb the data. Whereas clause shift is rotating syntactic trees to perturb data, the random shift is identical to a random swap in EDA \cite{taktasheva-etal-2021-shaking}. However, that paper focuses on evaluating the BERT and BART \cite{lewis-etal-2020-bart} attention layers for each token to observe their behavior but does not discuss their performance effects individually, nor do they disclose how the augmentation techniques are implemented.

Few research papers discuss controlled perturbations \cite{bayer2021survey} where pronoun tokens such as ``he" and ``she" are used to de-bias an NLP model \cite{zhao-etal-2018-gender}. This is a type of context-preserving augmentation technique. This has been done many times in the English language but has not been attempted within the Swedish language to the best of our knowledge. Moreover, our paper uses a data-driven approach to augment sentences in a controlled manner where any POS tag tokens can be specified.

\section{Proposed Method}
\subsection{Problem Statement}

Consider a low-resource setting, e.g., Swedish, where we only have limited data. The available dataset has low amounts of labeled data, and hence various augmentation tools are used to expand existing training data to make a better classification. However, one extra constraint is no dictionary synonyms are available. Our problem to solve is (1) to find a way to adapt the EDA-style augmentation for low-resource domains, (2) to measure how well EDDA performs on the Swedish datasets, (3) to examine how type specific similar work replacement (TSSR) affects classification.

\FloatBarrier

\subsection{Easy Distributional Data Augmentation (EDDA)}

\begin{figure}[!ht]
\centering
\includegraphics[width=1\columnwidth]{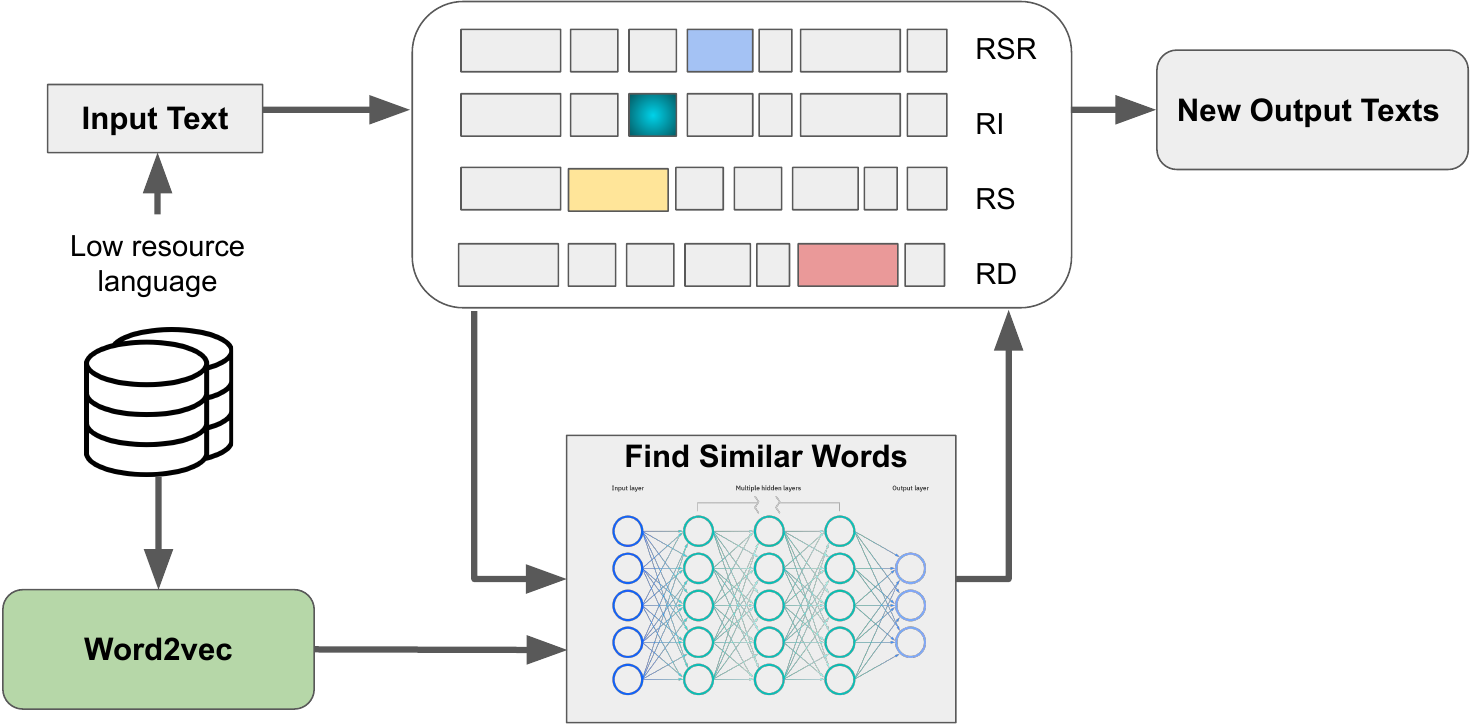} 
\caption{An overview of the EDDA framework.}
\label{fig:EDDA}
\end{figure}

We introduce EDDA, a novel technique to support text augmentation in low-resource languages (Figure \ref{fig:EDDA}). EDDA takes inspiration from EDA, which is a combination of many different augmentation methods \cite{wei-zou-2019-eda}. In this paper, we adapt the following strategies from EDA for low-resource data augmentation:

\begin{enumerate}
\item Random synonym replacement (RSR): We randomly select a small, user-defined fraction of words from the sentence, excluding stop words. A randomly chosen synonym replaces each replacement candidate.

\item Random insertion (RI): We randomly choose a small, user-defined fraction of positions within the sentence and insert a random synonym to a random word in the sentence.

\item Random swap (RS): We randomly choose a small, user-defined fraction of words and swap their positions.

\item Random deletion (RD): We randomly delete a small, user-defined fraction of words.
\end{enumerate}


With regards to embeddings, The name distributional in EDDA comes from distributional semantics which is capturing linguistic expressions as vectors that capture co-occurrence patterns in large corpora \cite{turney2010frequency,erk-pado-2008-structured}. This framework leverages this theory to augment various sentences using a language model like word2vec.


The synonym replacement is done, instead of using a lookup table, by using a word2vec model using its latent space to find the most similar word replacements. With the intuition that no functioning public synonym dictionary is available for Swedish, a Swedish word2vec model \cite{10.5555/2999792.2999959,10.2307/42636393} is used to generate word candidates with similar word distribution in an embedding space. Since it is not a dictionary that has a pure list of synonyms given a word, word2vec may not always find synonyms, but similar words that could occur in the same context. Thus, EDDA is a hybrid between a rule-based system such as EDA and a neural-based system.

While there are many distributional embeddings that could potentially be used, we use word2vec instead of e.g., BERT, since we use a particular token to find similar words in an embedding space which BERT's masked language modeling would not allow. Despite the fact that this might result in more randomness and potentially break the semantic meaning of a sentence, it allows us to support low-resource domains. Moreover, another advantage of using word2vec is that it still maintains the morphological coherence of suggested words compared to just using a synonym dictionary, such as SALDO \cite{Borin-Lars2013-188604}, as they are only in their base form. Additionally, the word2vec model is generally smaller than the BERT model, which may help with inference speed. The remainder of the EDA framework is used verbatim. Another benefit of using distributional semantic models to generate word replacement candidates is that non-strict synonyms (e.g., names or places) can be generated.

While word2vec is a well renowned algorithms where many languages have pre-trained models and even in cases where there is a lack of pre-trained models it is effortless to train such a model as it does not require any labeled data. Moreover, large scale language models like BERT may require heavy computational resources \cite{pmlr-v162-yao22c} to train, whereas word2vec may not, making it resource efficient. We claim that this adjustment can be of great benefit to language settings which lack good synonym dictionaries.

\subsection{ Type Specific Similar word Replacement (TSSR)}

\begin{figure}[!ht]
\centering
\includegraphics[width=1\columnwidth]{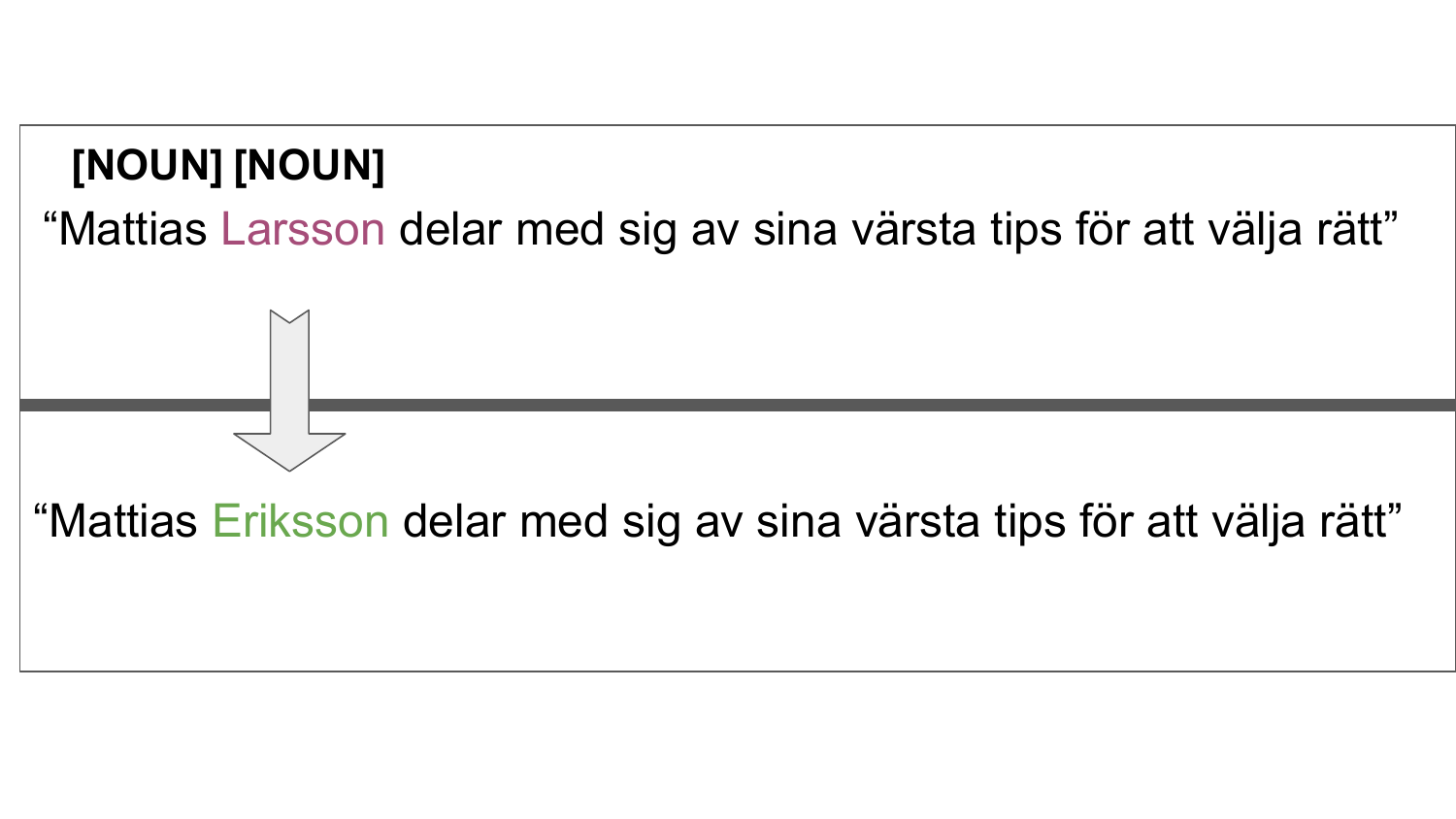} 
\caption{An example of TSSR replacing a noun word.}
\label{fig:tssr}
\end{figure}

When working with context-sensitive data, especially with sentiments, random synonym replacement might disrupt the semantic meaning of a sentence since the current EDA technique does not restrict replacing the word with synonyms from different types since it only looks at the list of synonyms in the dictionary. Therefore, we suggest constraining EDDA's synonym replacement by only replacing words with synonyms with the same POS tag, e.g., replacing verbs only with verb synonyms. Figure \ref{fig:tssr} shows one example where a noun token is chosen to be replaced and among two noun words `Larsson' is chosen and is replaced with `Eriksson'. 

To the best of our knowledge, no previous work has experimented with this method where word replacement using a language model (e.g., word2vec) with POS tag-specific perturbation has been done, especially within the low-resource domains. This allows for domain-specific augmentation that is more controllable and label-preserving in combination with EDDA.


\begin{algorithm}[!htb]
\DontPrintSemicolon
\SetKwInOut{Input}{Input}
\SetKwInOut{Output}{output}
\caption{TSSR pseudocode}
\label{alg:algorithm}
\Input{t: original text, s: token type, n: number of sentences to be created}
\KwResult{newSentences: list of new sentences}
newSentences = [] \;
\For{$i \leftarrow 1 \ldots n$} {
chosenToken = \textbf{FindRandomToken}(t, s)\;
candidateToken = \textbf{FindCandidate}(chosenToken)\;
newText = \textbf{Replace}(t, chosenToken, candidateToken)\;
newSentences.append(newText)\;
}
\KwRet{newSentences}

\end{algorithm}

\begin{figure}[!htb]
\centering
\includegraphics[width=1\columnwidth]{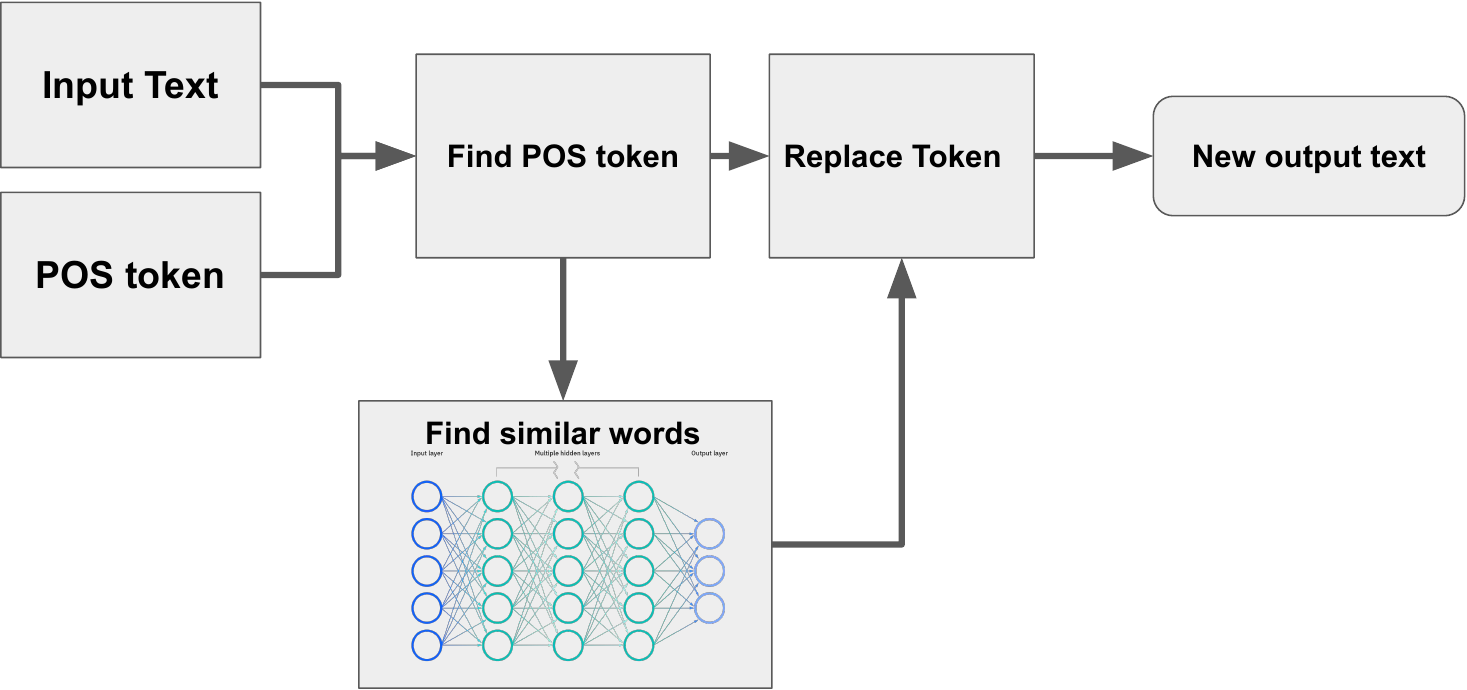} 
\caption{An overview of the TSSR framework.}
\label{fig:tssr2}
\end{figure}

The whole procedure of TSSR is depicted in Figure 3 and Algorithm 1. First, we iterate the process $n$ times to generate $n$ new sentences for each sentence where $n$ is a parameter (Algorithm 1, lines 1-2). A random token is chosen from the input text $t$ using the preferred token type $s$ as an input. A random POS token is selected if no token type is entered (line 3). After the chosen token, a new candidate token is generated using word embeddings (line 4) to replace the original text (lines 5). In the end, the new sentences are returned after the new altered sentences have been appended to the list (lines 6-7).

We acknowledge that POS taggers may not always be available in every low-resource language. The specified POS tags depend on the person or domain where the technique is being used to perturb the data. One thing to note about this technique is that it is not a pure synonym replacement but a similar word replacement based on the word embedding space, so it still does not depend on a synonym dictionary which low-resource languages might not have.

\section{Experimental Setup}

\begin{figure}[!htb]
\centering
\includegraphics[width=1\columnwidth]{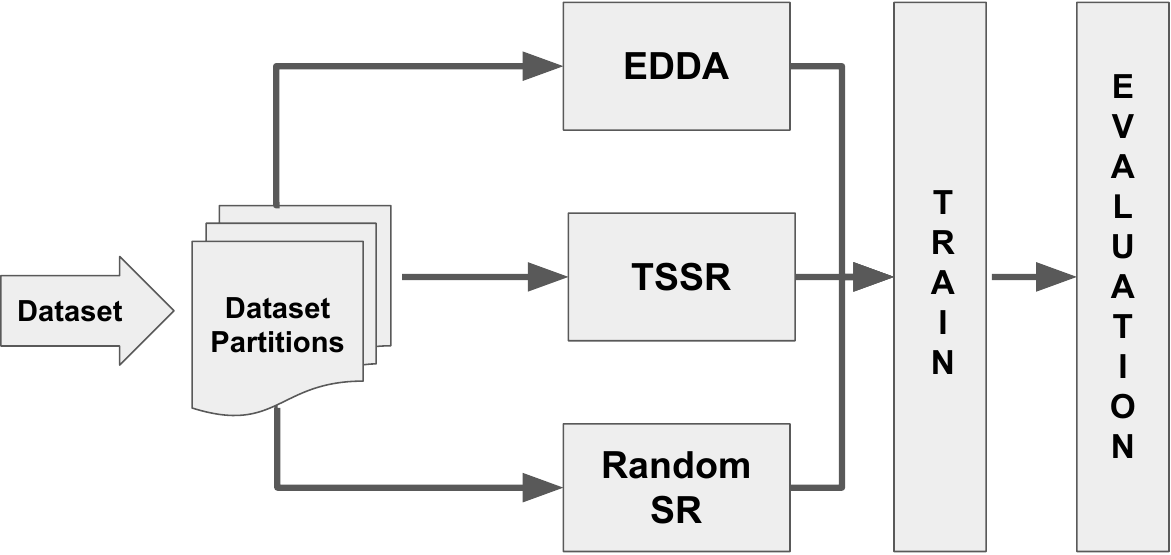} 
\caption{The augmentation pipeline for the experiments.}
\label{fig4}
\end{figure}

In this section, we describe our experiments on two downstream classification datasets to show how well the proposed
 text augmentation techniques (EDDA and TSSR) work in the Swedish language.

There are five main parts of the experiment, as also depicted in Figure \ref{fig4}, as follows:

\begin{enumerate}
\item Divide the dataset into multiple subpartitions.
\item With each dataset partition:
\begin{enumerate}
    \item Train baseline model with dataset partitions.
    \item Augment with EDDA and train another model.
    \item Augment with TSSR and train another model.
    \item Augment with RSR and train another model.
\end{enumerate}
\end{enumerate}

\subsection{Evaluation Methodology}
\subsubsection{Dataset Description}

The experiments are conducted with two publicly available datasets from a Swedish benchmarking dataset repository called SuperLim\footnote{\url{https://spraakbanken.gu.se/resurser/superlim}}. Since the datasets are already cleaned for research purposes, no special data cleaning or preprocessing is necessary \cite{Adesam-Yvonne2020-299130}. We use two datasets which represent the two most common problems in NLP, such as \textit{syntax analysis} and \textit{sentimental analysis}, as follows:

\begin{enumerate}
    \item \textbf{DALAJ}: A Swedish linguistic acceptability dataset \cite{volodina-etal-2021-dalaj}. This dataset contains a set of sentences where each sentence is denoted as linguistically correct or incorrect. The dataset has  predefined train, validation, and test splits with 7,682 training samples where 3,841 samples are classified into the correct group. On the other hand, the test set has 888 samples, where half are correct grammatical sentences and the other half are incorrect. The validation dataset is ignored in our experiment since our experiment does not have any parameter tuning. One thing to note about this dataset is when applying augmentation, only linguistically incorrect training samples are augmented, as augmenting the good samples has a higher chance of breaking the syntactic form of a sentence.
    \item \textbf{ABSA}: Aspect-based sentiment analysis (ABSA) is an annotated Swedish corpus for aspect-based sentiment analysis. This dataset includes various statements that are labeled from 1 (very negative) to 5 (very positive). The dataset also has  predefined train, validation, and test splits with 38,640 training samples and 4,830 test samples. Again, the validation dataset is ignored.
\end{enumerate}

\subsubsection{Baseline}
The baseline is trained with the training data with various subparts as described later. No augmentation techniques on the training data is applied to the baseline, only classification using the linear support vector machine (SVM) model \cite{10.1145/130385.130401} with BERT embeddings \cite{devlin-etal-2019-bert} is used. For all training attempts, SVM parameters used are the default parameters provided by Scikit-learn \cite{scikit-learn}.

\subsubsection{Experimental Settings}
 In this experiment, we use SVM, but any kind of linear classifier can be applied. The maximum limit is set to 512 tokens so that we can get all the information from the BERT model to classify our linear model. When extracting the BERT embeddings, only the [CLS] token are used after passing each text through the model. The BERT model is a 12-layer transformer with 768 hidden dimensions and 125M parameters \cite{swedish-bert}. Swedish is a low resource language but there was a Swedish Bert available hence it was used to generate embeddings. We acknowledge that such a large model may not exist in every low resource language but other feature extractors could be used instead.

\subsubsection{Implementation Details}
First, the training set is further split into 10,
20, 30, 40, 50, 60, 70, 80, 90, and 100 percent, where each partition is augmented and then added to that partition to train a linear model using BERT embeddings \cite{swedish-bert}. The reason behind making small stratified partitions is to re-create a scenario where insufficient data is available and to determine whether augmentation helps. Each dataset split is augmented using individual augmentation techniques, as shown in Figure \ref{fig4} where the augmentation is applied sequentially under the same partition to observe performance differences under controlled conditions. The perturbation rate is 20\% for every sentence that is augmented (i.e., 20\% of the tokens). Each sentence is augmented once using the described perturbation methods.
For the DALAJ dataset, only the incorrect samples are augmented using all the augmentation methods as they are already incorrect. Therefore, any perturbation is less likely to affect the class label. Moreover, each sentence is classified from the middle layer or layer six as it tends to have a high degree of syntax information \cite{rogers-etal-2020-primer} within the embedding before we pass it to the linear layer. 

On the other hand, for ABSA, all the augmentation techniques are applied regardless of class labels for ABSA as it is not as syntax sensitive in comparison to DALAJ. Moreover, the embeddings are extracted from the last layer, because we want to get the semantic embeddings from BERT for further experiments.

\subsubsection{Semantic Deviation}

After the augmentation has been applied, to observe the inner workings and impact on individual sentences, a check is done using similarity measures of non-augmented and augmented sentences
to assess the similarity. If an altered sentence largely deviates from its original form, this can be important to check, as very different sentences could destroy the semantics and could change the actual label. We use our $Deviction$ function to check the similarity between an original sentence $t$ and any augmented sentence $\hat{t}$ from $t$.
\[
    Deviction(t, \hat{t}) = 
\begin{cases}
    \text{``similar"},& \text{if } cos(t, \hat{t}) \geq \delta\\
    \text{``dissimilar"},              & \text{otherwise}
\end{cases}
\]
The deviation threshold $\delta$ is chosen at a level of 0.9 cosine similarity. Any sample below that is considered a different semantic sentence. The reason why 0.9 was used is because the augmented sentences should have high proximity to their original form which is important to preserve the sentiment label.  
The embedding is extracted using the same BERT model used for all the other experiments.

\section{Results}
This section is composed of two parts, where we showcase the F1 scores of various augmentation techniques on DALAJ \& ABSA. Lastly, we further investigate how much the sentiments deviate from one another for the ABSA dataset only. The techniques shown in the results are (1) baseline, (2) EDDA, (3) TSSR: controlled perturbation of selected parts of speech tags (in this case, nouns). (4) RSR: only using random synonym replacement. For a fair comparison, we use the exact data for each partition from the training set to augment and train models.

\subsection{DALAJ}

\begin{table}[ht!]
    \centering
    \begin{tabular}{l|cccc}
 \hline
Partition &  Baseline &  EDDA &  TSSR &  RSR \\
 \hline
       10\% &        0.56 &   \textbf{0.58} &         0.56 &         0.53 \\
       20\% &        0.55 &   0.60 &         0.59 &         \textbf{0.61} \\
       30\% &        0.58 &   \textbf{0.60} &         0.59 &         0.58 \\
       40\% &        0.56 &   0.61 &         0.57 &         \textbf{0.65} \\
       50\% &        0.56 &   0.61 &         0.61 &         \textbf{0.63} \\
       60\% &        0.60 &   0.61 &         0.61 &         \textbf{0.62} \\
       70\% &        \textbf{0.64} &   0.63 &         0.60 &         0.63 \\
       80\% &        \textbf{0.62} &   0.61 &         0.61 &         0.61 \\
       90\% &        0.55 &   \textbf{0.63} &         0.56 &         0.62 \\
      100\% &        \textbf{0.64} &   0.61 &         0.60 &         \textbf{0.64} \\
 \hline
\end{tabular}
     \caption{F1 scores on DALAJ under four different settings and ten different proportions of partitions.}
    \label{tab:tab1}
\end{table}

Table \ref{tab:tab1} shows overall F1 scores on DALAJ datasets under different settings. Up until 60\% partitions of data, all the augmentation technique improves classification on the DALAJ dataset. However, using more than 60\% of the data with augmentation tends to reduce the effectiveness of the said augmentation techniques. 

\begin{figure}[!htb]
\centering
\includegraphics[width=1\columnwidth]{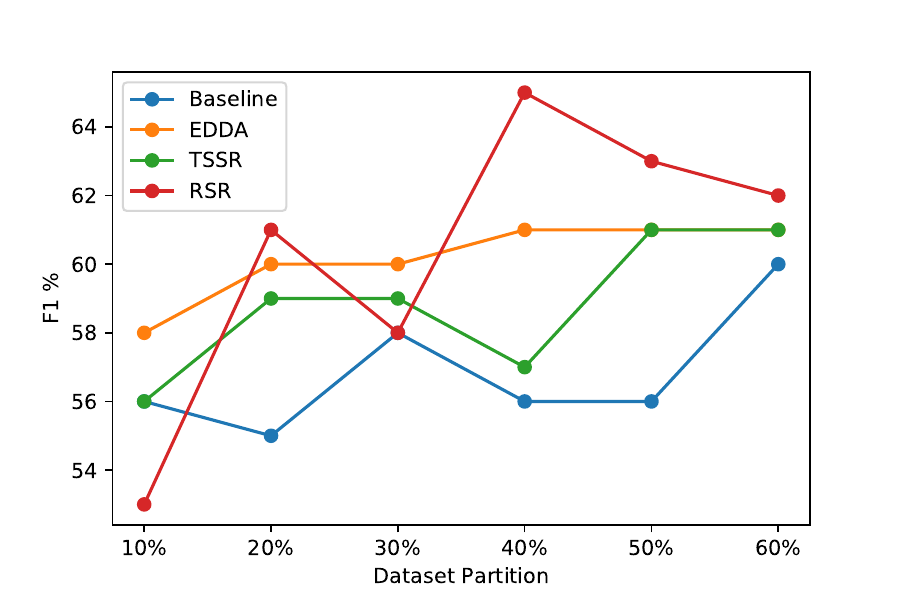} 
\caption{F1 scores on 10 to 60 percent of the DALAJ dataset under four different settings.}
\label{fig5}
\end{figure}
When comparing baseline to EDDA from 10\% to 60\% of the dataset, EDDA improves by 2.5\% on average. On the other hand, when comparing baseline to TSSR we get a 2\% average increase. But the best results appear under RSR compared to the baseline by a 3.5\% increase in classification performance.

One of the reasons for using text augmentation is when we have low amounts of data, we can use such techniques to improve models for downstream tasks. Figure \ref{fig5} supports that claim as by only using 40\% of the data, RSR improves by 9\% over the baseline, whereas EDDA improves F1 by 5\%. However, this is a case where TSSR only improves by 1\%.

The original paper, which has introduced DALAJ \cite{volodina-etal-2021-dalaj}, has also reported an F1 score of 62\% on the same test set. This is compared to our approach, where we only needed 40\% of the data to get 65\% on F1 using only RSR. Another proof of why augmentation can be effective with limited labeled data.

\subsubsection{EDDA \& RSR}
EDDA improves the performance in seven out of 10 partitions of the dataset, whereas RSR improves in six out of ten partitions. The augmentation works satisfactorily for this task, especially in low-data scenarios. Surprisingly, RSR performs exceptionally well, with only 40\% of the data overshadowing baseline with 100\% of the training data by 1\% in F1 score.

\subsubsection{TSSR}
This augmentation does improve in six out of 10 partitions. TSSR on this downstream task does not perform as well as EDDA and RSR. However, it consistently improves over the baseline, but it is not the most optimal augmentation technique to use in this classification dataset.

\subsection{ABSA}

\begin{table}[ht!]
\centering
\begin{tabular}{l|cccc}
\hline
Partition &  Baseline &  EDDA &  TSSR &  RSR \\
\hline

       10\% &        0.54 &    0.58 &    \textbf{0.59} &   0.53 \\
       20\% &        0.64 &    0.54 &    \textbf{0.67} &   0.64 \\
       30\% &        0.62 &    0.61 &    \textbf{0.64} &   0.63 \\
       40\% &        0.66 &    0.58 &    \textbf{0.69} &   0.59 \\
       50\% &        0.66 &    0.62 &    0.66 &   \textbf{0.69} \\
       60\% &        0.59 &    0.63 &    0.62 &   \textbf{0.65} \\
       70\% &        \textbf{0.70} &    0.65 &    0.67 &   0.66 \\
       80\% &        \textbf{0.71} &    0.66 &    0.73 &   0.63 \\
       90\% &        0.66 &    0.61 &    \textbf{0.72} &   0.66 \\
      100\% &        \textbf{0.74} &    0.63 &    0.71 &   0.67 \\
 \hline
\end{tabular}
    \caption{F1 scores on ABSA under four different settings and ten different proportions of partitions.}
    \label{tab:tab2}
\end{table}

\begin{figure}[!ht]
\centering
\includegraphics[width=1\columnwidth]{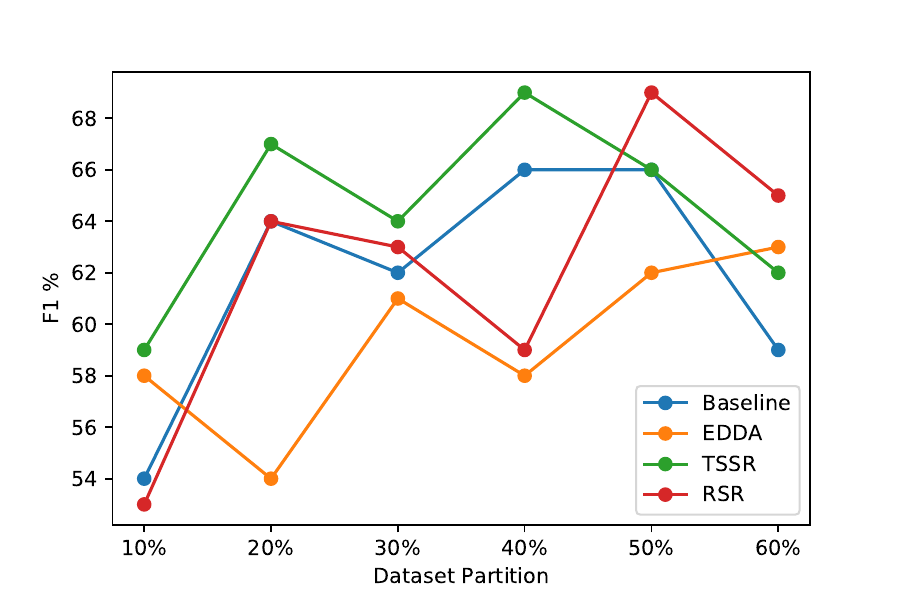} 
\caption{F1 scores on 10 to 60 percent of the ABSA dataset under four different settings.}
\label{fig6}
\end{figure}
Table \ref{tab:tab2} shows overall F1 scores on ABSA datasets under different settings. Until 60\% partitions of data, all the augmentation technique improves classification on the ABSA dataset. However, using more than 60\% of the data with augmentation tends to reduce the effectiveness of the said augmentation techniques. Figure \ref{fig6} shows that controlled perturbation on sentiment data improves the classification.

Comparing baseline to TSSR, we can see 2.7\% of F1 score increase on average. Additionally, using EDDA and RSR consistently improves F1 scores, albeit in some instances. When comparing EDDA with RSR, the performance seems to drop. RI, RD, and RS likely impact the classifications negatively, but as shown, RSR works slightly better than EDDA. 

\subsubsection{EDDA \& RSR}

EDDA does not consistently improve classification performance for this sentiment analysis task. It is to be noted that this is an aspect-based sentiment analysis dataset. Hence slightest perturbation could become detrimental to multi-label classification results.
EDA-style random perturbations are known to be bad for sentiment data as previous work has shown a decrease in classification scores \cite{10.1145/3366424.3383552, bayer2021survey,anaby2020not}. Another paper shows that random swap and deletion worsens sentence label preservation \cite{https://doi.org/10.48550/arxiv.2012.15466}. So our results coincide with previous work. However, RSR did produce good results a handful of times, therefore a justifiable augmentation technique that could work in a given dataset. 

\subsubsection{TSSR}

On eight out of ten partitions of this multi-labeled dataset, TSSR consistently improve classification performances. This is partly due to only changing certain noun token types that are more controlled. This gives a higher chance of not changing adverbs and adjectives, preserving the class label. Moreover, using a word2vec model to find replacements allows us to get a variety of words that may not be found in a standard dictionary, such as name replacements, e.g., Mattias Eriksson to Mattias Larsson.

\subsubsection{Semantic Deviation }

The semantic deviation is only assessed for the ABSA dataset to see how much augmentation affects the sentiment dataset. Only the sentiment dataset is used because it has the most chance of breaking when various augmentation techniques are applied.

\begin{table}[ht!]
\centering
 \begin{tabular}{l|c|c}

 \hline
 Technique &  Aug Sen below 0.9 &  \% Aug Sen below 0.9 \\
 \hline
    EDDA &  62,422 & 40.3\%  \\
    TSSR & 11,465 & 14.8\%  \\
 \hline
\end{tabular}
    \caption{Semantic deviation experiment on the ABSA dataset.}
    \label{tab:sd}
\end{table}
Table \ref{tab:sd} shows how many augmented sentences hold enough similarity to the original sentence. 40.3\% of the augmented sentences by EDDA do not meet our minimum criterion (i.e., cosine similarity 0.9) to be similar to the original sentence, while TSSR only produces 14.8\% of the synthetic sentences that have the similarity below 0.9, proving that TSSR preserves semantic proximity to the original sentence hence preserving the label. Therefore, it is safe to say TSSR can play a role as a competitive module together with EDDA in sentiment datasets.

\section{Conclusion}

We introduced EDDA, a modification of EDA without a huge dependency on language, and TSSR, a complementary method to EDDA, to replace synonyms given type specific information. We measured how these two techniques worked on the representative Swedish datasets and showed that those two techniques could improve DALAJ by 1\% over baseline with only 40\% of the training data. We also showed how well the presented augmentation worked with small amounts of labeled data and demonstrated that less data is most effective for augmentation to perform well. Moreover, augmentations may not always improve classification results but can still be very useful in most instances. We would like to emphasize that the techniques introduced in this paper are easily adaptable to other low-resource languages.
Our future work involves (1) testing the augmentation techniques in other low-resource languages, (2) testing on multiple different downstream tasks other than classifications, (3) extending the framework to other types of augmentation that is language agnostic or at least easily adaptable to any language.

\bibliography{aaai22.bib}

\end{document}